\documentclass[sigconf]{acmart}

\setcopyright{none}
\renewcommand\footnotetextcopyrightpermission[1]{} 
\AtBeginDocument{%
  }

\acmConference[MM '26]{ACM Multimedia 2026}{November 10--14, 2026}{Rio de Janeiro, Brazil}
\acmYear{2026}
\copyrightyear{2026}

\usepackage[utf8]{inputenc}
\usepackage{geometry}
\usepackage{booktabs} % 提供更好的水平线
\usepackage{multirow}
\usepackage{xcolor}   % 提供颜色支持
\usepackage{array}    % 优化表格列格式
\usepackage{caption}  % 优化标题设置
\usepackage{pifont} % 需要用到这个宏包提供图标
\usepackage{graphicx}
\usepackage{arydshln} % 用于 \cdashline 画虚线
\newcommand{\cmark}{\textcolor{green}{\ding{51}}} % 定义绿色的对号
\newcommand{\xmark}{\textcolor{red}{\ding{55}}}   % 定义红色的错号

\begin{document}

\title{MISID: A Multimodal Multi-turn Dataset for Complex Intent Recognition in Strategic Deception Games}

\author{Shufang Lin}
\authornote{Both authors contributed equally to this research.} % 定义共一脚注
\email{123090335@link.cuhk.edu.cn}
\affiliation{% 注意：这里必须补上第一作者的机构信息
  \institution{The Chinese University of Hong Kong, Shenzhen}
  \city{Shenzhen}
  \state{GuangDong}
  \country{China}
}

\author{Muyang Chen}
\authornotemark[1] % 打上和第一作者一样的星号（引用第1个authornote）
\email{123090028@link.cuhk.edu.cn}
\affiliation{%
  \institution{The Chinese University of Hong Kong, Shenzhen}
  \city{Shenzhen}
  \state{GuangDong}
  \country{China}
}

\author{Xiabing Zhou} % 在这里替换上导师的名字
\email{zhouxiabing@cnu.edu.cn}
\affiliation{%
  \institution{Capital Normal University}
  \city{Beijing}
  \country{China}
}

\author{Rongrong Zhang} % 在这里替换上导师的名字
\email{zhangrr@cnu.edu.cn}
\affiliation{%
  \institution{Capital Normal University}
  \city{Beijing}
  \country{China}
}

\author{Dayou Zhang} % 在这里替换上导师的名字
\authornote{Corresponding author.} % 这是第2个脚注，会自动分配不同的符号（如小匕首）
\email{zhangdayou@cnu.edu.cn}
\affiliation{%
  \institution{Capital Normal University}
  \city{Beijing}
  \country{China}
}

\author{Fangxin Wang} % 在这里替换上导师的名字
\email{wangfangxin@cuhk.edu.cn}
\affiliation{%
  \institution{The Chinese University of Hong Kong, Shenzhen}
  \city{Shenzhen}
  \state{GuangDong}
  \country{China}
}


\begin{abstract}
  Understanding human intent in complex multi-turn interactions remains a fundamental challenge in human-computer interaction and behavioral analysis. While existing intent recognition datasets focus mainly on single utterances or simple dialogues, real-world scenarios often involve sophisticated strategic interactions where participants must maintain complex deceptive narratives over extended periods. To address this gap, we introduce MISID, a comprehensive multimodal, multi-turn, and multi-participant benchmark for intent recognition. Sourced from high-stakes social strategy games, MISID features a fine-grained, two-tier multi-dimensional annotation scheme tailored for long-context discourse analysis and evidence-based causal tracking. Our systematic evaluation of state-of-the-art Multimodal Large Language Models (MLLMs) on MISID reveals critical deficiencies in complex scenarios, including text-prior visual hallucination, impaired cross-modal synergy, and limited capacity in chaining causal cues. Consequently, we propose FRACTAM as a baseline framework. Using a ``Decouple-Anchor-Reason'' paradigm, FRACTAM reduces text bias by extracting pure unimodal factual representations, employs two-stage retrieval for long-range factual anchoring, and constructs explicit cross-modal evidence chains. Extensive experiments demonstrate that FRACTAM enhances mainstream models' performance in complex strategic tasks, improving hidden intent detection and inference while maintaining robust perceptual accuracy. Our dataset is available at \url{https://naislab.cn/datasets/MISID}.
\end{abstract}



\begin{teaserfigure}
  \centering
  \vspace{-8pt}
  \includegraphics[width=0.9\textwidth]{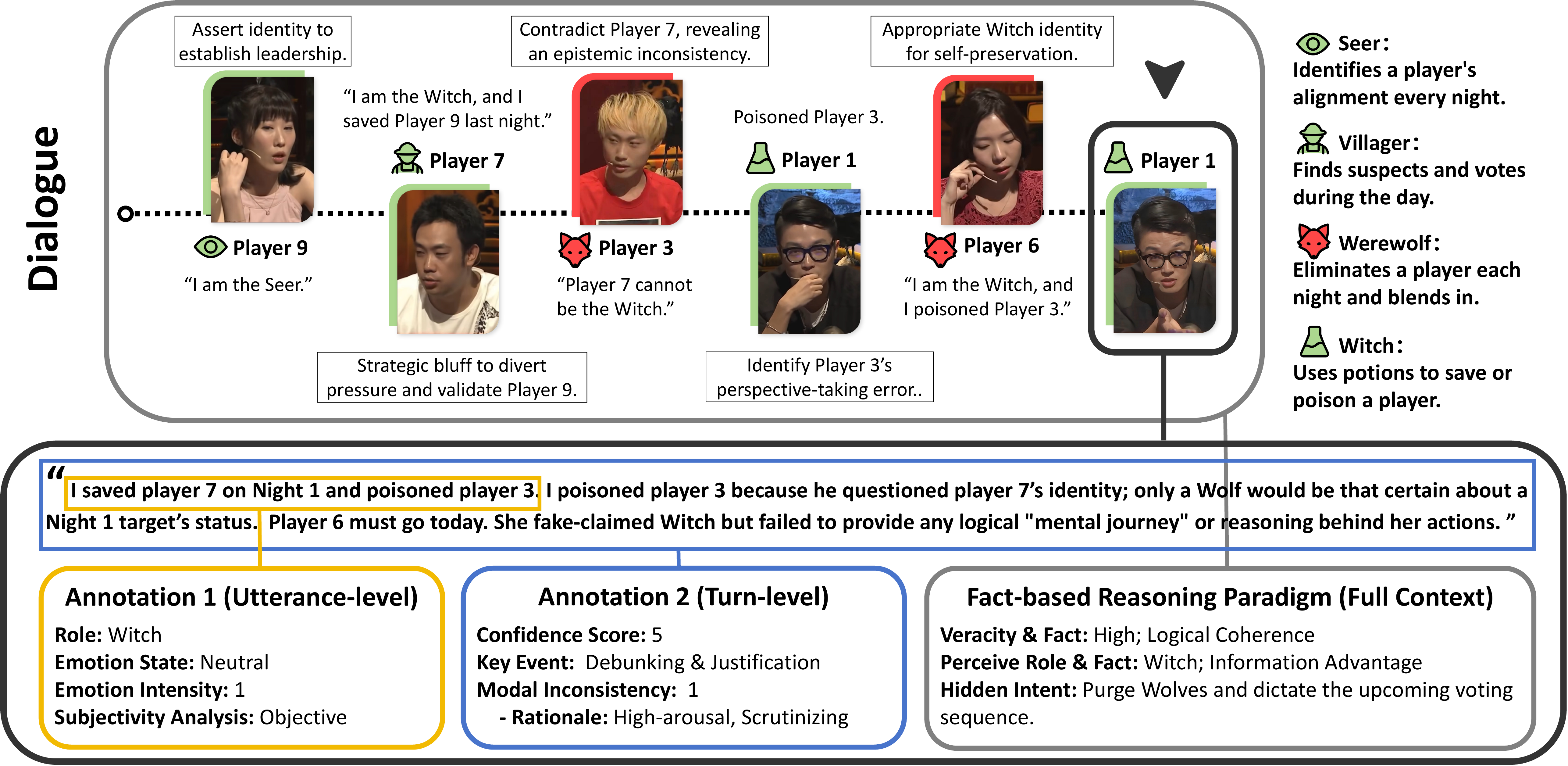} 
  \vspace{-6pt}
  \caption{An overview of the MISID benchmark. (Top) A multi-participant strategic dialogue timeline exhibiting hidden tactics. (Bottom) Our multi-dimensional annotation scheme and fact-based reasoning paradigm for deducing hidden intents.}
  %\vspace{-3pt}
  \Description{Top: A multi-turn interaction timeline where participants engage in deceptive strategies, such as identity theft and bluffing, to mask their true alignments. Bottom: The multi-dimensional annotation scheme of MISID. It systematically progresses from utterance-level micro-states (e.g., emotion, subjectivity) and turn-level discourse analysis (e.g., modal inconsistency, key events) to a comprehensive causal reasoning chain, guiding models to decipher hidden intents from long-contextual clues.}
  \label{fig:double_column_image}
\end{teaserfigure}


\maketitle

\section{Introduction}

\textit{``Real life consists of bluffing, of little tactics of deception, of asking yourself what is the other man going to think I mean to do.''} \\
\null\hfill --- John von Neumann

Human social interaction is fundamentally a game of incomplete information \cite{spence1978job, harsanyi1967games}. From high-stakes political negotiations to everyday workplace dynamics, people strategically conceal their true intentions behind euphemisms \cite{pinker2008logic, brown1987politeness}, leverage multimodal signals to deceive or persuade, and continuously adapt their strategies across prolonged exchanges \cite{buller1996interpersonal}. While current Artificial Intelligence (AI) excels at processing transparent, explicit instructions, it falters when confronted with this defining feature of human sociality: the pervasive gap between what people say and what they truly mean \cite{mahowald2024dissociating, sap2022neural}. As AI advances toward genuine social intelligence \cite{DBLP:conf/iclr/Zhou0MZYQMBFNS24}, the benchmark for evaluating model capabilities can no longer rely on surface semantic comprehension \cite{bisk2020experience}. Instead, the critical challenge lies in penetrating strategic expressions to perceive hidden intentions, tracking how these intentions evolve over long-term interactions, and capturing the subtle cross-modal leaks, such as a fleeting micro-expression belying verbal agreement, that expose genuine psychological states.

To endow AI with such sophisticated social intelligence, comprehensive benchmarks are imperative. However, existing intent recognition datasets exhibit severe limitations when addressing dynamic and covert interactions \cite{liu2022research, shen2026hidden, sarathy2020reasoning}. Most benchmarks are predominantly ``transparent'' and static, focusing on instantaneous, single-sentence interactions where speech perfectly aligns with thought \cite{zhang2022mintrec, zhang2024mintrec2}. By stripping away the strategic concealment and long-term causal chains prevalent in real-world settings, these datasets force models to over-rely on superficial textual signals ~\cite{geirhos2020shortcut}. Even the few datasets that address intent concealing often remain at the level of binary judgments or shallow ``stimulus-response'' patterns \cite{la2025beyond}. They lack fine-grained analysis of intent dynamics, hidden motivations, and the complex historical facts that drive dynamic interactions, fundamentally hindering models from developing fact-grounded reasoning capabilities \cite{DBLP:conf/iclr/Zhou0MZYQMBFNS24}.

To bridge this gap, we introduce MISID---a multi-turn, multimodal, and multi-participant intent recognition benchmark. Constructed from multimodal sources of high-pressure social strategy games involving deception, reasoning, and voting-based elimination, MISID naturally recreates realistic strategic scenarios with extreme information density. Featuring 3,962 high-quality utterance segments with  temporally aligned audiovisual modalities, the dataset transcends traditional single-dimensional labeling by introducing a two-tier multi-dimensional annotation scheme. It scales from utterance-level micro-states (sentiment, intensity) to macro-level, long-range multimodal discourse analysis. By explicitly annotating cross-modal inconsistencies, cross-turn logical contradictions, and precisely localized historical key facts, MISID creates an evidence-based causal reasoning paradigm. It shifts the learning objective from superficial guessing to tracking complex derivation chains based on hard evidence.

To gauge the capability of current models in such strategic scenarios, we systematically evaluated mainstream Large Language Models (LLMs) alongside their multimodal counterparts (MLLMs) on MISID. We observed that the concealment and temporal dynamics of human intentions uniquely captured by MISID pose fundamental challenges to existing multimodal architectures. Specifically, their failures are characterized by three primary dilemmas: (1) \textit{Text-prior Visual Hallucination}, where models force visual representations to align with dominant textual logic rather than relying on objective visual input; (2) \textit{Causal Threading Limitations}, where models struggle to penetrate noisy, multi-turn contexts to connect scattered historical events into a coherent causal chain; and (3) \textit{Impaired Modal Synergy}, where distributing causal chains across modalities paradoxically increases fact-related error rates compared to unimodal settings.

Addressing these multi-turn multimodal reasoning dilemmas, we propose the FRACTAM (\textbf{F}act-grounded \textbf{R}easoning \textbf{A}nd \textbf{C}ausal \textbf{T}hreading \textbf{A}cross \textbf{M}odalities) framework. Based on a structured ``decoupling-anchoring-reasoning'' paradigm, FRACTAM directly tackles the observed model failures. First, it extracts unimodal symbolic representations from audiovisual signals to mitigate textual prior dominance. Second, it employs a dual-stage, long-range fact anchoring mechanism to isolate causal variables from historical noise accurately. Finally, by constructing cross-modal causal chains, it effectively repairs impaired modality synergy, enabling the model to perform robust, evidence-based reasoning in complex gaming scenarios.

Our primary contributions are as follows:
\begin{itemize}
\item We propose MISID, the first multi-turn, multimodal, and multi-dimensionally labeled benchmark dataset designed within a complex, high-pressure strategic environment.
\item We comprehensively benchmark state-of-the-art unimodal and multimodal models, exposing their critical bottlenecks (e.g., text-prior hallucination and impaired modal synergy) in multi-turn hidden intent recognition.
\item We introduce the FRACTAM framework, a novel structured reasoning paradigm that effectively overcomes current limitations in long-range cross-modal causal threading, establishing a strong baseline for future research in artificial social intelligence.
\end{itemize}

\begin{table}[t]
\caption{\textbf{Comparison of MISID with existing multimodal dialogue and intent recognition benchmarks.} Depth: Literal (Explicit) or Underlying (Implicit) Intentions. CSE: Complex Strategic Environment. FCA: Fact-based Causal Annotation. Length: Dialogue Turn Ranges.}
\vspace{-6pt}
\centering
\small
\renewcommand{\arraystretch}{1.1} % 稍微增加行高，更接近原图
\setlength{\tabcolsep}{2pt}    % 调整列间距以适应页面
\resizebox{\columnwidth}{!}{%
\begin{tabular}{l | l | c c c | c c | c }
\hline
\textbf{Dataset} & \textbf{Depth} & \textbf{Text} & \textbf{Audio} & \textbf{Video} & \textbf{CSE} & \textbf{FCA} & \textbf{Length} \\ \hline

% 1. Explicit emotion alignment and multi-dimensional intent classification
MCIC \cite{yuan2022mcic} & Explicit & \cmark & \xmark & \xmark & \xmark & \xmark & 10-30 \\

MSAIRS \cite{shi2025impact} & Explicit & \cmark & \xmark & \xmark & \xmark & \xmark & 1-10 \\

SLURP \cite{bastianelli2020slurp} & Explicit & \cmark & \cmark & \xmark & \xmark & \xmark & 1-10 \\

FSC \cite{qian2021speech} & Explicit & \cmark & \cmark & \xmark & \xmark & \xmark & 1-10 \\

MINDS-14 \cite{gerz2021multilingual} & Explicit & \cmark & \cmark & \xmark & \xmark & \xmark & 1-10 \\

MIntRec \cite{zhang2022mintrec} & Explicit & \cmark & \cmark & \cmark & \xmark & \xmark & 1-10 \\

MIntRec 2.0 \cite{zhang2024mintrec2} & Explicit & \cmark & \cmark & \cmark & \xmark & \xmark & 10-20 \\

EMOTyDA \cite{saha2020towards} &  Explicit  & \cmark & \cmark & \cmark & \xmark & \xmark & 1-10 \\

EmoInt-MD \cite{singh2022emoint} &  Explicit  & \cmark & \cmark & \cmark & \xmark & \xmark & 1-10 \\

MC-EIU \cite{liu2024emotion} & Explicit & \cmark & \cmark & \cmark & \xmark & \xmark & 1-10 \\

BID \cite{maharana2022multimodal} & Explicit & \cmark & \cmark & \cmark & \xmark & \xmark & 1-10 \\

MECPE \cite{wang2022multimodal} & Explicit & \cmark & \cmark & \cmark & \xmark & \cmark & 10-40 \\

Genesis \cite{li2025genesis} & Explicit & \cmark & \cmark & \cmark & \xmark & \cmark & 100-500 \\

% 2. Hidden psychological states, deception and masking behavior
Open Domain \cite{perez2015experiments}& Implicit& \cmark & \xmark & \xmark & \xmark & \xmark & 1-10 \\

CSC \cite{hirschberg2005distinguishing}& Implicit & \xmark & \cmark & \xmark & \xmark & \xmark & 1-10 \\

Bag-of-Lies \cite{gupta2019bag} & Implicit & \cmark & \cmark & \cmark & \xmark & \xmark & 1-10 \\

Box of Lies \cite{soldner2019box}& Implicit & \cmark & \cmark & \cmark & \xmark & \xmark & 10-20 \\

Real Life Trials \cite{perez2015deception}& Implicit & \cmark & \cmark & \cmark & \xmark & \xmark & 1-10 \\

MELD \cite{poria2019meld}& Implicit & \cmark & \cmark & \cmark & \xmark & \xmark & 10-100 \\

MUStARD \cite{castro2019towards} & Implicit & \cmark & \cmark & \cmark & \xmark & \xmark & 1-10 \\

% 3. LLM-driven multimodal domain (Complex Reasoning / Non-literal / Implicit Intent)
MultiMET \cite{zhang2021multimet} & Implicit & \cmark & \xmark & \xmark & \xmark & \xmark & 1-10 \\

MDID \cite{kruk2019integrating} & Implicit & \cmark & \xmark & \xmark & \xmark & \xmark & 1-10 \\

CraigslistBargain \cite{he2018decoupling} & Implicit & \cmark & \xmark & \xmark & \cmark & \xmark & 10-30 \\

IntentQA \cite{li2023intentqa} & Implicit & \cmark & \cmark & \cmark & \xmark & \cmark & 1-10 \\ 

Diplomacy \cite{niculae2015linguistic}& Implicit & \cmark & \xmark & \xmark & \cmark & \xmark & 100-600 \\ \hline

% OURS
MISID (OURS) & Implicit& \cmark & \cmark & \cmark & \cmark & \cmark & 154-555\\ \hline
\end{tabular}}

%\Description{A table comparing the proposed MISID dataset with 14 existing benchmarks (e.g., IEMOCAP, MIntRec, MUStARD, Diplomacy) across several dimensions. The columns indicate the Core Task, availability of Text, Audio, and Video modalities, presence of a Complex Strategic Environment (CSE), presence of Fact-based Causal Annotation (FCA), and Context Length measured in utterances.

%The data illustrates that most existing datasets focus on basic Emotion or Explicit Intent, typically feature short conversational lengths (mostly 1 to 20 utterances), and lack either CSE, CFT, or comprehensive multimodal support. In stark contrast, the bottom row highlights the proposed MISID dataset. MISID focuses on "Hidden Intent", includes all three modalities (Text, Audio, Video), provides both CSE and FCA, and features exceptionally long contexts ranging from 154 to 555 utterances, marking it as the only comprehensive dataset across all listed dimensions.}
\vspace{-12pt}
\end{table}

\section{Related Work}

\textbf{Dialogue Understanding and Intent Recognition Datasets.} 
Recent advancements in affective computing and dialogue understanding depend heavily on multimodal benchmarks \cite{wu2025multimodal}. Mainstream datasets (e.g., SLURP \cite{bastianelli2020slurp}, FSC \cite{qian2021speech}, MIntRec \cite{zhang2022mintrec} and MELD \cite{poria2019meld}) primarily target explicit emotion alignment and multi-dimensional intent classification. To capture hidden psychological states, various datasets have been developed to study deception and masking behaviors (e.g., Real-Life Trial Data \cite{perez2015deception}, Bag-of-Lies \cite{gupta2019bag}, and MUStARD \cite{castro2019towards}), investigating phenomena from physiological leakage in constrained environments to intentional masking in the wild. Furthermore, recent works have begun extending into the LLM-driven multimodal domain (e.g., MECPE \cite{wang2022multimodal}, IntentQA \cite{li2023intentqa}, Genesis \cite{li2025genesis}). However, these existing datasets have notable limitations: they primarily assume explicit expressions, are confined to highly restricted contexts with single-dimensional features, and often restrict annotations to static, binary judgments. Consequently, they fail to track the dynamic evolution of intents grounded in key facts during real-world strategic concealment. To fill this gap, MISID provides hierarchical annotations, multi-turn multimodal synchronized data, and a fact-based reasoning paradigm.

\textbf{Intent and Deception Detection Methods.} 
To capture temporal cues and multi-party interactions, traditional methods typically employ sequence models, basic attention mechanisms and Graph Neural Networks (e.g., LSTMs \cite{hochreiter1997long}, TCNs \cite{bai2018empirical}, DialogueGCN \cite{ghosal2019dialoguegcn}, DAG-ERC \cite{shen2021directed}). Advanced deep networks, such as CTNet \cite{lian2021ctnet}, further integrate global contexts through deep attention frameworks. More recently, Large Language Models (LLMs) have been heavily leveraged to provide powerful zero-shot reasoning capabilities in multimodal social scenarios \cite{hurst2024gpt, comanici2025gemini, xai2025grok41, qwen3.5_2026}. However, these methodologies present critical limitations in high-pressure, long-range social games. Traditional sequence and graph-based models struggle to maintain long-range strategic contexts. They are prone to capturing superficial temporal correlations rather than robust causal dependencies, and often suffer from feature over-smoothing that obscures sparse triggers. Meanwhile, despite their strong reasoning potential, LLMs are hampered by long-term memory decay, modality bias, and a lack of rigorous causal frameworks to constrain hallucinations. To overcome these bottlenecks, our proposed FRACTAM framework integrates context-aware key-fact extraction, multimodal perception, and fact-based rigorous reasoning to effectively address these challenges.

\begin{table}[t]

\caption{\textbf{Dataset Statistics.} Physical statistics (left) and multimodal annotation distributions (right) of the MISID dataset. VAD: Voice Activity Detection. Inconsistency: Cross-modal Incongruence During Interactions.}
\vspace{-6pt}
\centering
\resizebox{\columnwidth}{!}{%
    \begin{tabular}{l l @{\hspace{2em}} l l}
    \toprule
    % --- 表头 ---
    \multicolumn{2}{l}{\textbf{Dimension}} & \multicolumn{2}{l}{\textbf{Annotation Distribution}} \\
    \midrule

    % --- 左侧：原有维度数据 | 右侧：新统计的标注数据 ---
    Total Participants & 15 & Subjectivity: Subj & 3,259 \\
    Avg. Participants & 12 & Subjectivity: Obj & 703 \\
    Role Instances & 120 & Emotion: Calm & 2802 \\
    Total Utterances & 3,962 & Emotion: Others & 1160 \\
    Total Duration & 9.15 hours & Deception: Truth & 2191 \\
    Max Turns & 555 & Deception: Lie & 1771 \\
    Avg. Turns & 374.7 & Inconsistency: High & 1542  \\
    Mean VAD Ratio & 92.8\% & Inconsistency: Low & 2420 \\

    \bottomrule
    \end{tabular}%
}
%\Description{A two-column statistical table detailing the scale and annotation distribution of the MISID dataset. 

%The left column, titled "Dimension," highlights the dataset's physical properties. It comprises 15 total participants, 120 role instances, and 3,962 total utterances, amounting to a total duration of 9.15 hours. The conversations are notably lengthy, featuring a maximum of 555 turns, an average of 374.7 turns, and a high mean Voice Activity Detection (VAD) ratio of 92.8%.

%The right column, titled "Annotation Distribution," breaks down the count of labeled instances across four categories. For Subjectivity, there are 3,259 subjective and 703 objective instances. Emotion annotations consist of 2,802 "Calm" and 1,160 "Others" labels. The Deception distribution is relatively balanced, with 2,191 "Truth" and 1,771 "Lie" instances. Finally, modal Inconsistency is categorized into 1,542 "High" and 2,420 "Low" instances.}
\label{tab:dataset_stats}
\vspace{-12pt}
\end{table}

\section{Dataset}

\noindent \textbf{Dataset Overview} \
We introduce MISID, a multi-turn multimodal dataset based on complex strategic games. The dataset contains 3,962 high-quality utterance segments, covering 120 role-specific utterance instances from 15 participants, with detailed statistics and annotations summarized in Table 2. Each segment includes video and audio modalities obtained from public online videos and precisely synchronized to ensure temporal alignment.

\noindent \textbf{Data Processing} \
Audio tracks were uniformly standardized to 16kHz mono PCM format. Speaker diarization was conducted with Pyannote 3.1 \cite{bredin2023pyannote, plaquet2023powerset}, augmented by multi-dimensional filtering, heuristic merging and manual correction. Simultaneously, the visual modality utilized face detection and DBSCAN-based identity clustering techniques \cite{ester1996density} to extract individual sequences, ensuring strict temporal alignment with the audio components.

\noindent \textbf{Data Annotation} \
MISID adopts a two-tier multi-dimensional annotation scheme: the first layer records foundational background (identity, basic emotion, intensity, and subjectivity); the second layer targets discourse analysis (anchoring key events, confidence, and modality inconsistency), establishing cross-modal and multi-turn factual reasoning chains. The annotation workflow follows a two-stage pattern: initial labels are first generated by Large Language Models, followed by rigorous manual verification by four human annotators, integrating iterative peer reviews and cross-modal consistency audits.

\noindent \textbf{Fact-based Reasoning Paradigm} \
We constructed an evidence-based causal reasoning paradigm aimed at guiding models to learn factual derivation chains. By leveraging the two-tier annotations and ground truth, this paradigm precisely locates key contextual cues to guide the model in reconstructing logical chains and inferring participants' deceptive behaviors and hidden intentions, serving as a standard benchmark for model training and evaluation.

\begin{figure}[t] % 标准 figure，放在单栏顶部
  \centering
  % width=1.0\columnwidth 确保它占满单栏宽度
  \includegraphics[width=1\columnwidth]{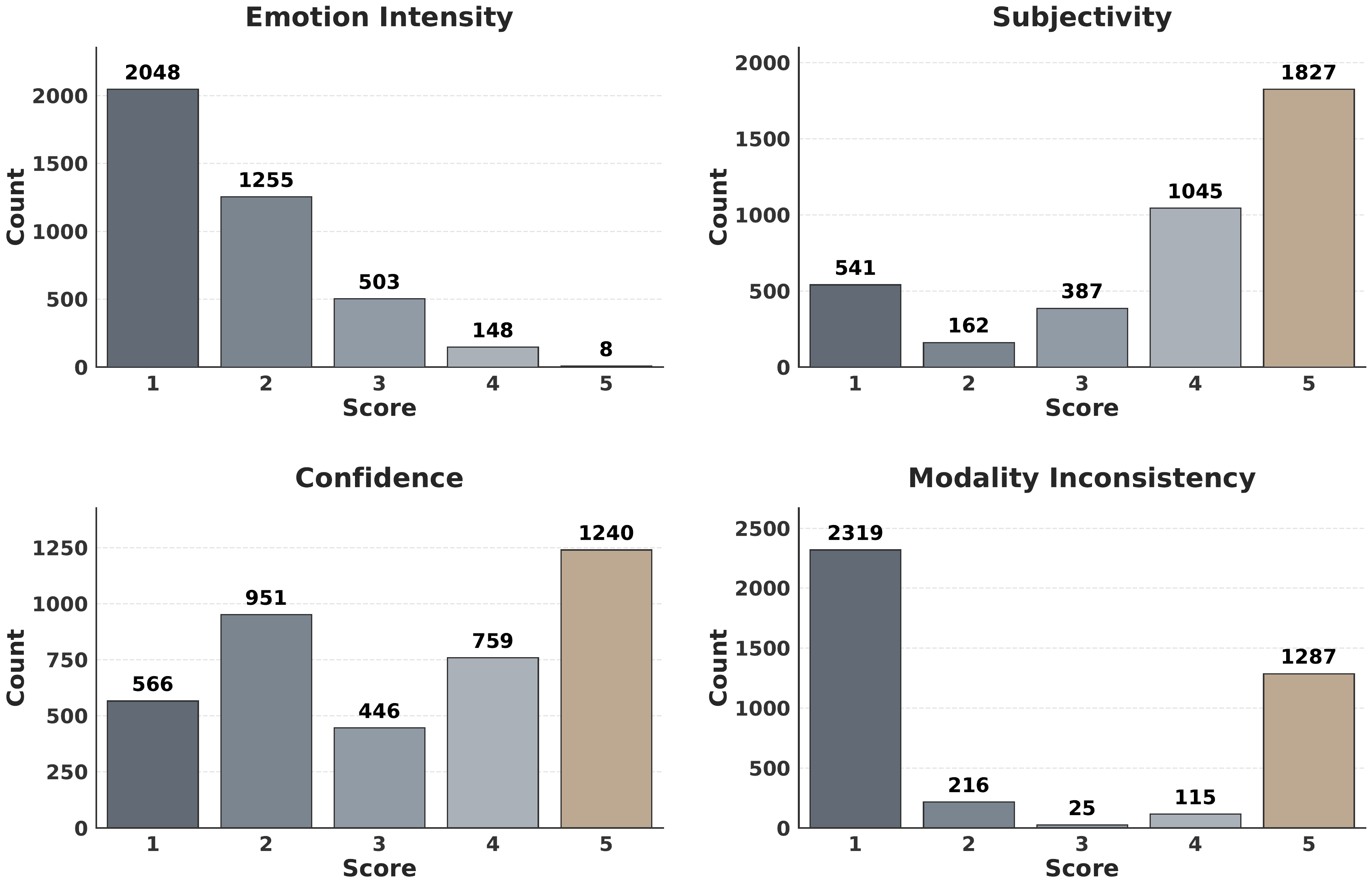}
  \caption{Distribution of multi-dimensional annotations in the MISID dataset. The x-axis denotes the annotation score, and the y-axis represents the occurrence frequency.}
  \Description{Description:
  A set of four bar charts presented in a 2x2 grid, illustrating the statistical distribution of annotations. For all charts, the vertical axis represents count and the horizontal axis ranges from 1 to 5.

  The top-left chart, Emotion Intensity, shows a strong right-skewed or L-shaped distribution, where score 1 has the highest count (approx. 2000) and counts decrease significantly for subsequent scores.

  The top-right chart, Subjectivity, depicts a complex distribution with peaks at 1 (approx. 500) and 5 (approx. 1800), with low counts for scores 2, 3, and 4, indicating an overall increase after an initial moderate count.

  The bottom-left chart, Confidence, displays a complex, multimodal shape with prominent peaks at score 2 (approx. 1000) and particularly score 5 (approx. 1200), while scores 1, 3, and 4 have lower values.

  The bottom-right chart, Modality Inconsistency, features an L-shaped distribution with a dominant peak at score 1 (approx. 2300), followed by very low counts for scores 2, 3, and 4, and a secondary moderate peak at score 5 (approx. 1300).}
  \label{fig:single_arch}
\end{figure}

\section{Methodology}

FRACTAM is designed to address intent concealment and multimodal signal conflicts in complex social games. Its core logic is to first achieve the objective decoupling of physical facts, then anchor historical evidence via long-range retrieval, and ultimately conduct evidence-based reasoning through the construction of cross-modal causal chains.

\subsection{Modal Decoupling and Text Reconstruction}

To eliminate modality feature contamination induced by textual logic alignment, FRACTAM utilizes Multimodal Large Language Models (MLLMs) to implement strict unimodal fact decoupling. For the input signal $\mathcal{X}_i = \{x_i^t, x_i^v, x_i^a\}$ at the $i$-th interaction round, we disable early cross-modal attention and independently input the non-textual modalities $x_i^m$ into the MLLM $\Theta$. By introducing a restrictive prompt $\mathbf{p}_m$, the model is constrained to output only objective descriptions of physical states. This decoding process can be abstracted as the maximization of the sequence joint probability under given conditions:
\begin{equation}
f_i^m = \mathop{\arg\max}_{\mathbf{y}} P_{\Theta} \big( \mathbf{y} \mid \Phi_m(x_i^m), \mathbf{p}_m \big), \quad m \in \{v, a\}
\end{equation}
where the $\Phi_m$ represents the modality-specific projection mapping. Through Equation (1), both visual and audio signals are decoded into factual texts $f_i^v$ and $f_i^a$, respectively. Ultimately, the multimodal signals are decoupled and uniformly constructed into a unified plain-text fact set $\mathcal{F}_i = \{x_i^t, f_i^v, f_i^a\}$.

\subsection{Hybrid Long-range Fact Anchoring}

Addressing the challenge of sparse causal variables in long-term interactions, FRACTAM introduces a two-stage long-range fact anchoring mechanism. Given the current fact set $\mathcal{F}_t$ and the historical memory bank $\mathcal{H} = \{\mathcal{F}_i\}_{i=1}^{t-1}$, we first conduct a dual-path recall in the lexical (lex) and semantic (sem) feature spaces, calculating the initial relevance via weighted Reciprocal Rank Fusion \cite{cormack2009reciprocal}:
\begin{equation}
\Omega(\mathcal{F}_i, \mathcal{F}_t) = \sum_{\rho \in \{\text{lex}, \text{sem}\}} \frac{\omega_\rho}{\eta + \pi_\rho(\mathcal{F}_i, \mathcal{F}_t)}
\end{equation}
where $\pi_\rho(\cdot, \cdot)$ is the descending ranking function in a specific retrieval space, and $\eta$ is a smoothing term. Based on this score, we extract the top M facts with the highest $\Omega$ scores from $\mathcal{H}$ to form a high-confidence candidate subset $\mathcal{H}_{cand}$:
\begin{equation}
\mathcal{H}_{cand} = \operatorname*{Top-M} \Big\{ \mathcal{F}_i \in \mathcal{H} \ \Big|\ \Omega(\mathcal{F}_i, \mathcal{F}_t) \Big\}
\end{equation}
To further capture deep contextual dependencies, we introduce a cross-encoder (Qwen3-Reranker \cite{zhang2025qwen3}) $\Phi_{ce}$ to re-rank the candidate set. For each historical fact $\mathcal{F}_j$ in the candidate set, the model calculates its deep semantic relevance score $s_{ce}$ with the current fact $\mathcal{F}_t$ through a joint attention mechanism:
\begin{equation}
s_{ce}(\mathcal{F}_t, \mathcal{F}_j) = \sigma \Big( \mathbf{W}_{ce}^\top \, \Phi_{ce} \big( [\mathcal{F}_t ; \mathcal{F}_j] \big) \Big)
\end{equation}
where $[\cdot ; \cdot]$ denotes sequence concatenation, $\mathbf{W}_{ce} \in \mathbb{R}^{d_{ce} \times 1}$ is the linear projection weight, and $\sigma$ is the activation function. Based on this score, we select the top K most relevant historical facts from $\mathcal{H}_{cand}$ to construct the final retrieval context bank $\mathcal{C}_{ret}$:
\begin{equation}
\mathcal{C}_{ret} = \operatorname*{Top-K} \Big\{ \mathcal{F}_j \in \mathcal{H}_{cand} \ \Big|\ s_{ce}(\mathcal{F}_t, \mathcal{F}_j) \Big\}
\end{equation}

\begin{figure}[t]
  \centering
  % 第一张图，宽度设为 0.48 确保留有微量间隙
  \includegraphics[width=0.49\columnwidth]{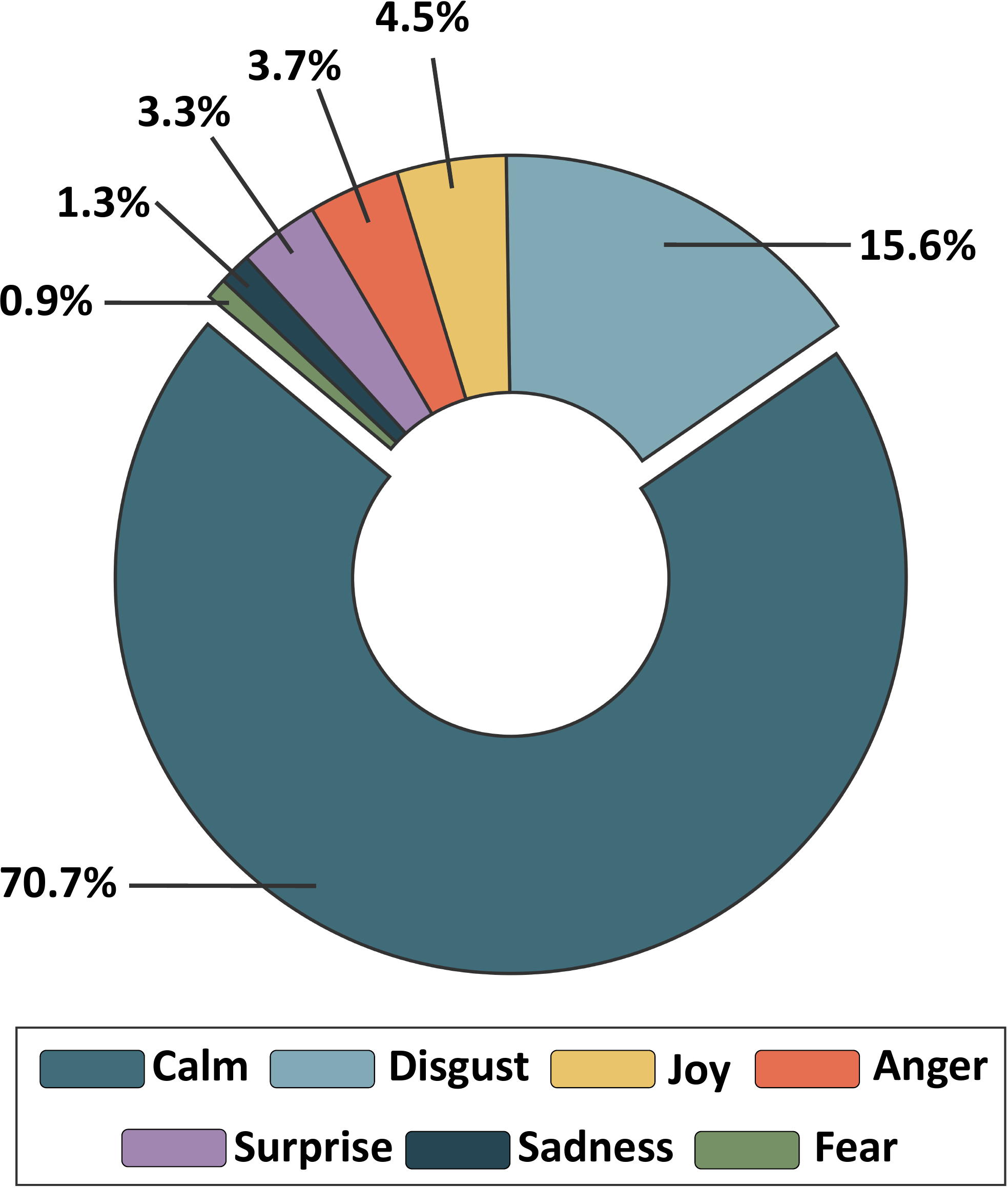}
  \hfill % 这个命令会自动撑满中间的空格，把两张图推向两端
  \includegraphics[width=0.49\columnwidth]{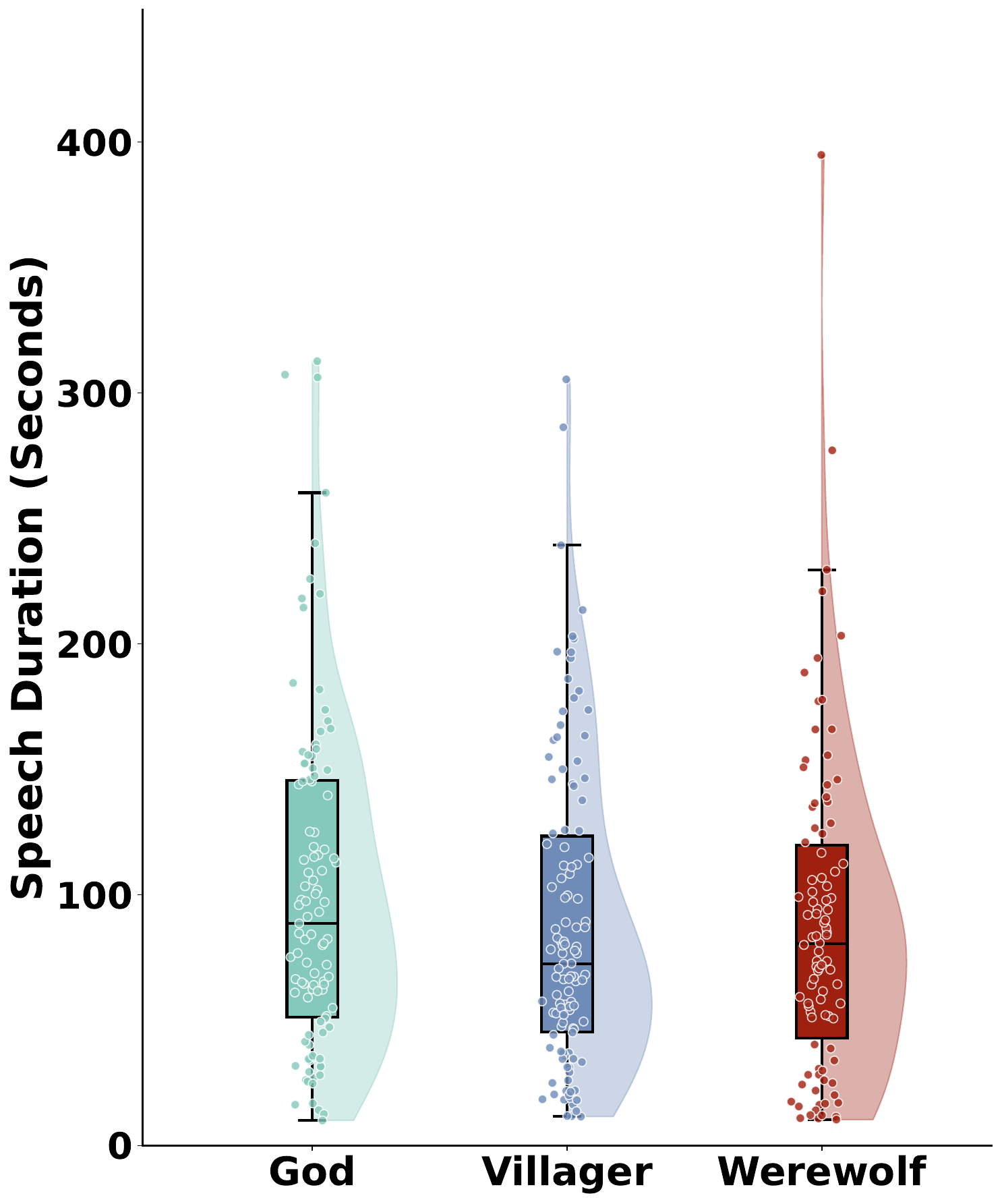}
  
  \caption{Emotion category distribution (left) and speech duration distribution (right) in MISID dataset.}
  \Description{Figure 4 contains two subplots. The left subplot is a donut chart showing the distribution of emotion categories, heavily dominated by "Calm" (70.7\%), followed by "Disgust" (15.6\%), and minor percentages for Joy, Anger, Surprise, Sadness, and Fear. The right subplot is a radar chart comparing the average performance of VLLM, LLM, and FRACTAM across eight metrics (ESA, EIS, SA, IJA, IRS, LDA, LRS, HIS). The red polygon representing FRACTAM encompasses the largest area, indicating superior performance across all evaluation dimensions compared to the baselines.}
  \label{fig:side_by_side}
\end{figure}

\begin{figure*}[t] % 标准 figure，放在单栏顶部
  \centering
  % width=1.0\columnwidth 确保它占满单栏宽度
  \includegraphics[width=\textwidth]{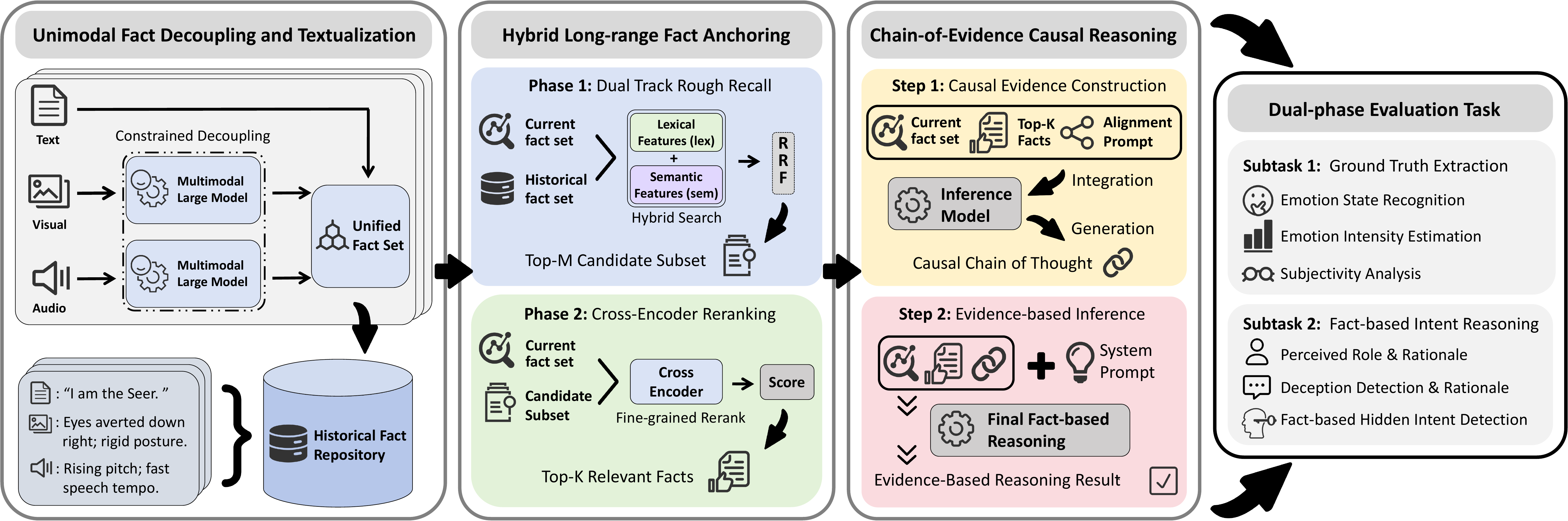}
  \caption{Overall architecture of the FRACTAM framework. The pipeline standardizes multimodal inputs into objective text, retrieves historical evidence via dual-stage hybrid search, and constructs explicit logical chains to deduce underlying intents for dual-phase evaluation.}
  \Description{A flowchart illustrating the FRACTAM framework, divided into four sequential stages from left to right.

  Unimodal Fact Decoupling and Textualization: Text, Visual, and Audio signals are independently processed by Multimodal Large Models under constrained decoupling to form a Unified Fact Set, which is then archived in a Historical Fact Repository.

  Hybrid Long-range Fact Anchoring: A two-phase retrieval process. Phase 1 performs a Dual Track Rough Recall combining lexical and semantic features via Reciprocal Rank Fusion (RRF) to generate a Top-M candidate subset. Phase 2 uses a Cross-Encoder to rerank this subset into Top-K relevant facts.

  Chain-of-Evidence Causal Reasoning: The current fact set and Top-K retrieved facts prompt an Inference Model to construct a Causal Chain of Thought (Step 1). This chain, guided by a system prompt, drives the final evidence-based reasoning result (Step 2).

  Dual-phase Evaluation Task: The final outputs address Subtask 1 (Ground Truth Extraction for emotion and subjectivity) and Subtask 2 (Fact-based Intent Reasoning for role perception, deception detection, and hidden intent).}
  \label{fig:single_arch_2}
\end{figure*}

\subsection{Causal Threading and Intent Inference}

To resolve the ``modality synergy degradation'' caused by black-box fusion, we introduce explicit modality fact evidence chains. Based on the current modality-decoupled fact set $\mathcal{F}_t$ and the retrieval context $\mathcal{C}_{ret}$, and under the constraint of the cross-modal alignment prompt $\mathbf{p}_{align}$, the reasoning model $\Psi$ is forced to construct an explicit causal evidence chain $\mathcal{T}_c$:
\begin{equation}
\mathcal{T}_c = \mathop{\arg\max}_{\mathbf{z}} P_{\Psi} \big( \mathbf{z} \mid \mathcal{F}_t, \mathcal{C}_{ret}, \mathbf{p}_{align} \big)
\end{equation}
Subsequently, under the logical hard constraint of the system inference prompt $\mathbf{p}_{sys}$, the model $\Psi$ must strictly follow the evidence chain $\mathcal{T}_c$ and key facts to generate the final output:
\begin{equation}
\mathbf{\hat{Y}} = \mathop{\arg\max}_{\mathbf{y}} P_{\Psi} \big( \mathbf{y} \mid \mathcal{T}_c, \mathcal{F}_t, \mathcal{C}_{ret}, \mathbf{p}_{sys} \big)
\end{equation}
Through this proposed paradigm, the model $\Psi$ effectively coordinates various modalities with an interpretable evidence chain, yielding the evidence-based reasoning result $\mathbf{\hat{Y}}$ regarding fact determination and intent analysis threads.

\section{Evaluation Metrics}

For ground truth extraction, we adopt standard metrics along with a distance-decay scoring mechanism based on numerical gaps. For hidden intent extraction involving complex causal reasoning, $n$-gram metrics based on lexical overlap and cosine similarity calculations are no longer sufficient. Therefore, we employ the \textbf{LLM-as-a-Judge} evaluation paradigm, utilizing a Large Language Model evaluator $\mathcal{E}_{LLM}$ for quantification.

\subsection{State Recognition Metrics}

State recognition metrics are used to evaluate the model's precision in determining the key game states of participants and serve as factual grounding for downstream reasoning metrics. Among these, \textbf{Role Accuracy (RA)} measures the model's judgment regarding the participants' specific identities and their respective camps. We utilize a tiered scoring method:

\begin{equation}
RA = \frac{1}{N} \sum_{i=1}^N \mathcal{F}_{role}(y_i^{role}, \hat{y}_i^{role}), \quad
\end{equation}

where $\mathcal{F}_{role}$ is the tiered scoring function, which assigns a full score for a completely correct identity, a partial score if only the camp is correct, and 0 for an incorrect judgment.

\textbf{Deception Binary Accuracy (DBA)} evaluates the model's classification performance regarding ``whether a lie was told'':

\begin{equation}
DBA = \frac{100}{N} \sum_{i=1}^N I(y_i^{lie} = \hat{y}_i^{lie}), \quad
\end{equation}

where $I(\cdot)$ is the indicator function, which equals 1 when the prediction matches the ground truth and 0 otherwise.

\subsection{Evidence-based Reasoning Metrics}

Evidence-based reasoning metrics are used to evaluate the model's fact recall and logical argumentation capabilities. We quantify the model evaluation results across three dimensions: Identity Reasoning (IRS), Lie Details (LDS), and Hidden Intent (HIS). For any reasoning task $task \in \{IRS, LDS, HIS\}$, the evaluator $\mathcal{E}_{LLM}$ assesses the fact-grounded responsiveness ($\Phi_{FG}$) and logical consistency ($\Phi_{LC}$) of the predicted reasoning $\hat{R}$ using the following formula:

\begin{equation}
Score_{task} = \frac{1}{N} \sum_{i=1}^N \Big( \alpha \cdot \Phi_{FG}(F_i^*, \hat{R}_i) + \beta \cdot \Phi_{LC}(R_i^*, \hat{R}_i) \Big), \quad
\end{equation}

where $F_i^*$ represents the key facts, $R_i^*$ represents the standard logic, and $\alpha, \beta$ are balancing weights. Furthermore, to suppress fact-detached logical hallucinations, a pre-state hard truncation mechanism is introduced:

\begin{equation}
HIS_i = \min\Big( \tau, \ HIS_i \Big), \quad \text{if } RA_i = 0 \lor DBA_i = 0, \quad
\end{equation}

where $\tau$ is the penalty threshold. This mechanism ensures that when the ground truth determination is incorrect, the score for its corresponding reasoning explanation is constrained.

\begin{table*}[t]
\caption{Performance of LLMs on various tasks. The scores are reported in percentage (\%). The \colorbox{pink}{best} and the \colorbox{yellow}{second best} results are denoted by pink and yellow. LLM implies the exclusive use of the text modality. FRACTAM refers to LLMs assisted by the FRACTAM framework.}
\label{tab:performance}
\centering
\renewcommand{\arraystretch}{1.1} % 稍微增加行间距以适应上下标
\resizebox{\textwidth}{!}{
\begin{tabular}{c|lccc|cccccc}
\toprule
% \multicolumn{1}{c|}{} & Model & \shortstack{Emotion State\\Accuracy} & \shortstack{Emotion\\Intensity Score} & \shortstack{Subjectivity\\Accuracy} & \shortstack{Identity Judgment\\Accuracy} & \shortstack{Identity\\Reasoning Score} & \shortstack{Lie Detection\\Accuracy} & \shortstack{Lie Reasoning\\Score} & \shortstack{Hidden Intent\\Inference Score} \\
{} & Model & \shortstack{Emotion State\\Accuracy} & \shortstack{Emotion\\Intensity Score} & \shortstack{Subjectivity\\Accuracy} & \shortstack{Identity Judgment\\Accuracy} & \shortstack{Identity\\Reasoning Score} & \shortstack{Lie Detection\\Accuracy} & \shortstack{Lie Reasoning\\Score} & \shortstack{Hidden Intent\\Inference Score} \\
% \multicolumn{1}{c|}{} & GPT-o1 [21] & 65.51 & 50.16 & 44.03 & 30.07 & & & & & \\
\midrule
\multirow{10}{*}{VideoLLM}
& GPT-4o \cite{hurst2024gpt} & 76.43 & 73.75 & 77.14 & 46.15 & \colorbox{yellow}{43.77} & 50.16 & \colorbox{yellow}{45.77} & \colorbox{yellow}{39.77} \\
& Grok-4-Fast \cite{xai2025grok41} & \colorbox{yellow}{86.21} & 79.89 & 90.80 & \colorbox{yellow}{53.37} & 41.88 & \colorbox{yellow}{53.82} & 39.46 & 32.69 \\
& Qwen3-VL-235B-A22B \cite{bai2025qwen3} & 79.05 & 76.19 & 87.14 & 28.85 & 31.69 & 42.31 & 31.46 & 21.69 \\
& Qwen3-VL-Flash & 69.23 & 46.70 & 87.91 & 46.15 & 41.54 & 46.15 & 41.58 & 28.27 \\
& Qwen3-VL-Max & 74.05 & 75.57 & 91.22 & 30.77 & 34.85 & 46.15 & 35.54 & 23.42 \\
& Qwen3-VL-Plus & 72.83 & 61.78 & 90.58 & 15.38 & 27.69 & 34.62 & 34.77 & 23.27 \\
& Qwen3.5-Plus \cite{qwen3.5_2026} & 83.78 & \colorbox{yellow}{80.28} & \colorbox{pink}{94.37} & 13.46 & 26.65 & 30.77 & 29.65 & 22.12 \\
& Gemini-2.5-flash \cite{comanici2025gemini} & 82.55 & 78.55 & 91.64 & 21.15 & 31.08 & 50.16 & 36.42 & 26.85 \\
& Gemini-3-flash \cite{google2025gemini3flash} & \colorbox{pink}{87.09} & \colorbox{pink}{85.10} & \colorbox{yellow}{94.04} & \colorbox{pink}{53.85} & \colorbox{pink}{52.73} & \colorbox{pink}{53.88} & \colorbox{pink}{57.58} & \colorbox{pink}{49.52} \\
\cdashline{2-10}
& Average & 79.02 & 73.09 & 89.43 & 34.35 & 36.88 & 45.34 & 39.14 & 29.73 \\
\midrule
\multirow{11}{*}{LLM}
& Claude-Sonnet-4.5 \cite{anthropic2025claude45}& \colorbox{pink}{89.01} & \colorbox{yellow}{81.87} & 82.42 & 50.48 & 45.12 & \colorbox{yellow}{57.69} & \colorbox{yellow}{52.38} & \colorbox{yellow}{43.35} \\
& DeepSeek-R1 \cite{guo2025deepseek} & 83.45 & 78.10 & 83.79 & \colorbox{pink}{65.38} & \colorbox{pink}{56.54} & 42.31 & 43.85 & 36.35 \\
& DeepSeek-V3 \cite{liu2024deepseek}& 66.91 & 71.40 & 80.94 & 51.92 & \colorbox{yellow}{49.04} & 50.33 & 49.52 & \colorbox{pink}{44.23} \\
& GLM-4.7 \cite{zhipu2025glm47}& 83.79 & 81.03 & 92.07 & 42.31 & 45.38 & 34.62 & 38.27 & 31.96 \\
& GPT-4o & 80.28 & 75.70 & 81.34 & 51.92 & 44.81 & 49.84 & 48.27 & 41.23 \\
& GPT-5.1 \cite{openai2025gpt51}& 82.73 & 74.32 & \colorbox{yellow}{92.73} & 44.23 & 48.54 & 53.85 & \colorbox{pink}{53.15} & 36.46 \\
& Grok-4-Fast & 86.59 & 81.10 & 90.24 & \colorbox{yellow}{57.69} & 47.62 & \colorbox{pink}{61.54} & 50.65 & 40.85 \\
& MiniMax-M2.1 \cite{li2025minimax}& 80.08 & \colorbox{pink}{82.27} & 86.06 & 42.31 & 40.69 & 23.08 & 27.31 & 21.69 \\
& Qwen3-Max \cite{yang2025qwen3}& 77.18 & 77.52 & 87.58 & 26.92 & 30.15 & 34.62 & 29.04 & 22.85 \\
& Qwen3.5-Plus & \colorbox{yellow}{87.32} & 81.08 & \colorbox{pink}{93.24} & 17.31 & 27.50 & 34.62 & 32.65 & 23.27 \\
\cdashline{2-10}
& Average & 81.73 & 78.44 & 87.04 & 45.05 & 43.54 & 44.25 & 42.51 & 34.22 \\
\midrule
\multirow{11}{*}{\textbf{FRACTAM}}
& Claude-Sonnet-4.5 & \colorbox{pink}{90.75$_{\uparrow 1.74}$} & 81.46$_{\downarrow 0.41}$ & 82.88$_{\uparrow 0.46}$ & \colorbox{yellow}{61.96$_{\uparrow 11.48}$} & 56.53$_{\uparrow 11.41}$ & \colorbox{pink}{61.14$_{\uparrow 3.45}$} & \colorbox{yellow}{57.27$_{\uparrow 4.89}$} & \colorbox{pink}{53.27$_{\uparrow 9.92}$} \\
& DeepSeek-R1 & 83.73$_{\uparrow 0.28}$ & 80.18$_{\uparrow 2.08}$ & 85.66$_{\uparrow 1.87}$ & \colorbox{pink}{72.22$_{\uparrow 6.84}$} & \colorbox{pink}{63.87$_{\uparrow 7.33}$} & 50.25$_{\uparrow 7.94}$ & 50.86$_{\uparrow 7.01}$ & 45.59$_{\uparrow 9.24}$ \\
& DeepSeek-V3 & 69.53$_{\uparrow 2.62}$ & 71.2$_{\downarrow 0.20}$ & 81.92$_{\uparrow 0.98}$ & 61.69$_{\uparrow 9.77}$ & 57.83$_{\uparrow 8.79}$ & 56.6$_{\uparrow 6.27}$ & 55.05$_{\uparrow 5.53}$ & 48.02$_{\uparrow 3.79}$ \\
& GLM-4.7 & 83.39$_{\downarrow 0.40}$ & 81.3$_{\uparrow 0.27}$ & 93.34$_{\uparrow 1.27}$ & 55.01$_{\uparrow 12.70}$ & 55.0$_{\uparrow 9.62}$ & 42.56$_{\uparrow 7.94}$ & 45.97$_{\uparrow 7.70}$ & 43.04$_{\uparrow 11.08}$ \\
& GPT-4o & 79.87$_{\downarrow 0.41}$ & 75.9$_{\uparrow 0.20}$ & 83.11$_{\uparrow 1.77}$ & 54.85$_{\uparrow 2.93}$ & 50.16$_{\uparrow 5.35}$ & 53.85$_{\uparrow 4.01}$ & 51.61$_{\uparrow 3.34}$ & 46.09$_{\uparrow 4.86}$ \\
& GPT-5.1 & 84.14$_{\uparrow 1.41}$ & 74.59$_{\uparrow 0.27}$ & \colorbox{yellow}{94.29$_{\uparrow 1.56}$} & 56.93$_{\uparrow 12.70}$ & \colorbox{yellow}{58.41$_{\uparrow 9.87}$} & \colorbox{yellow}{60.36$_{\uparrow 6.51}$} & \colorbox{pink}{59.38$_{\uparrow 6.23}$} & \colorbox{yellow}{52.83$_{\uparrow 16.37}$} \\
& Grok-4-Fast & \colorbox{yellow}{88.92$_{\uparrow 2.33}$} & 80.62$_{\downarrow 0.48}$ & 92.56$_{\uparrow 2.32}$ & 55.56$_{\downarrow 2.13}$ & 49.75$_{\uparrow 2.13}$ & 59.8$_{\downarrow 1.74}$ & 45.1$_{\downarrow 5.55}$ & 40.01$_{\downarrow 0.84}$ \\
& MiniMax-M2.1 & 82.02$_{\uparrow 1.94}$ & \colorbox{yellow}{82.96$_{\uparrow 0.69}$} & 86.1$_{\uparrow 0.04}$ & 43.29$_{\uparrow 0.98}$ & 43.13$_{\uparrow 2.44}$ & 31.59$_{\uparrow 8.51}$ & 32.58$_{\uparrow 5.27}$ & 29.72$_{\uparrow 8.03}$ \\
& Qwen3-Max & 80.03$_{\uparrow 2.85}$ & 78.2$_{\uparrow 0.68}$ & 87.4$_{\downarrow 0.18}$ & 33.76$_{\uparrow 6.84}$ & 38.65$_{\uparrow 8.50}$ & 41.09$_{\uparrow 6.47}$ & 35.85$_{\uparrow 6.81}$ & 30.44$_{\uparrow 7.59}$ \\
& Qwen3.5-Plus & 87.16$_{\downarrow 0.16}$ & \colorbox{pink}{83.55$_{\uparrow 2.47}$} & \colorbox{pink}{94.85$_{\uparrow 1.61}$} & 37.82$_{\uparrow 20.51}$ & 44.7$_{\uparrow 17.20}$ & 41.7$_{\uparrow 7.08}$ & 39.23$_{\uparrow 6.58}$ & 38.84$_{\uparrow 15.57}$ \\
\cdashline{2-10}
& Average & 82.95$_{\uparrow 1.22}$ & 79.0$_{\uparrow 0.56}$ & 88.21$_{\uparrow 1.17}$ & 53.31$_{\uparrow 8.26}$ & 51.8$_{\uparrow 8.26}$ & 49.89$_{\uparrow 5.64}$ & 47.29$_{\uparrow 4.78}$ & 42.79$_{\uparrow 8.57}$ \\
\bottomrule
\end{tabular}
}
\end{table*}

\section{Experiments}

\subsection{Experimental Setup}

We benchmark a diverse suite of mainstream models. These are categorized into VideoLLMs and Text-only LLMs. All models were evaluated in a zero-shot setting. For text-only LLMs, visual dynamics were represented through textual behavioral descriptions. 

\subsection{Model Limitations Analysis}

Our empirical analysis identifies three critical bottlenecks hindering current foundation models in the MISID environment:

\textbf{Text-prior Visual Hallucination.} In deception scenarios, people often exhibit a distinct discrepancy between their verbal language and non-verbal cues. We observe that VideoLLMs heavily over-rely on the audio transcript. When textual claims conflict with visual evidence, they frequently ignores authentic visual cues and generating hallucinated visual descriptions that spuriously align with the deceptive text. Consequently, the average Hidden Intent Inference Score for VideoLLMs languishes at a mere 29.73\%.

\textbf{Limitations in Threading Causal Clues.} Both VideoLLMs and text-only LLMs demonstrated severe limitations in temporal causal reasoning. Their reasoning scores consistently trail their binary judgment accuracies. For instance, while text-only LLMs average 44.25\% in Lie Detection Accuracy, their Lie Reasoning Score drops to 42.51\%, indicating that even when models guess the correct label, their underlying causal justification is often flawed.

\textbf{Impaired Modal Synergy.} The integration of raw visual modalities degrades, rather than enhances, higher-order strategic reasoning. As shown in Table \ref{tab:performance}, text-only LLMs consistently outperform VideoLLMs across all strategic metrics. Specifically, the average Identity Judgment Accuracy for text-only LLMs is 45.05\%, compared to a significantly lower 34.35\% for VideoLLMs, which indicates compromised modal synergy. Rather than complementing, the unrefined visual modality acts as noise and distracts the model from logical deduction.

\section{Conclusion}

In this work, we address the limitations of existing intent recognition benchmarks by introducing MISID, a pioneering multi-turn, multimodal, and multi-participant dataset constructed from high-pressure strategic games. By offering precise audiovisual alignment and a novel multi-dimensional annotation scheme that captures cross-modal inconsistencies, MISID shifts the evaluation paradigm toward complex, fact-grounded causal reasoning. Our comprehensive evaluation of state-of-the-art models on MISID reveals critical bottlenecks in current architectures, notably text-prior visual hallucination and impaired modal synergy during complex social interactions. To establish a robust baseline, we propose the FRACTAM framework, which mitigates these multi-turn reasoning dilemmas through strict modality decoupling and explicit evidence chain construction. Ultimately, by capturing the pervasive gap between surface expressions and hidden psychological states, we envision MISID serving as a vital catalyst for the multimedia community, driving future research toward more interpretable, robust, and genuinely socially intelligent multimodal models.

\bibliographystyle{ACM-Reference-Format}
\bibliography{sample-base}

\end{document}